\documentclass{ecai} 

\usepackage{latexsym}
\usepackage{amssymb}
\usepackage{amsmath}
\usepackage{amsthm}
\usepackage{booktabs}
\usepackage{enumitem}
\usepackage{graphicx}
\usepackage{color}

\usepackage{comment}
\usepackage{subcaption}

\newtheorem{definition}{Definition}

\newcommand{\idest}{{\it i.e.}}

\newcommand{\etal}{{\it et al.}}

\DeclareMathOperator{\cost}{cost}
\DeclareMathOperator{\pre}{pre}
\DeclareMathOperator{\eff}{eff}
\DeclareMathOperator{\positive}{pos}
\DeclareMathOperator{\negative}{neg}
\providecommand\condition{\ensuremath{c}}

\providecommand\tuple[1]{\ensuremath\langle#1\rangle}

\DeclareMathOperator{\fluents}{\mathcal{F}}
\DeclareMathOperator{\actions}{\mathcal{A}}
\DeclareMathOperator{\fluent}{f}
\DeclareMathOperator{\action}{a}

\DeclareMathOperator{\planningtask}{\Pi}
\DeclareMathOperator{\planningdomain}{\Xi}
\DeclareMathOperator{\transition}{\gamma}

\DeclareMathOperator{\plan}{\pi}
\DeclareMathOperator{\optimalplan}{\pi^{*}}

\providecommand\initialstate{\ensuremath{\mathcal{I}}}
\providecommand\goalcondition{\ensuremath{G}}

\providecommand{\datapipeline}{\ensuremath{\mathcal{D}}}

\providecommand{\graphoperators}{\ensuremath{\mathcal{O}}}
\providecommand{\graphoperator}{\ensuremath{O}}
\providecommand{\graphoperatorid}{\ensuremath{\mathit{id}}}

\providecommand{\graphedges}{\ensuremath{\mathcal{E}}}
\providecommand{\graphedge}{\ensuremath{E}}
	
\providecommand{\graphgroups}{\ensuremath{\mathcal{G}}}
\providecommand{\graphgroup}{\ensuremath{G}}
\providecommand{\groupoperators}[1]{\ensuremath{O_{\graphgroup_{#1}}}}

\providecommand{\operatortags}{\ensuremath{T}}
\providecommand{\graphtags}{\ensuremath{\mathcal{T}}}
\providecommand{\graphtag}{\ensuremath{t}}

\providecommand{\graphimages}{\ensuremath{\mathcal{I}}}
\providecommand{\graphimage}{\ensuremath{I}}

\providecommand{\graphsdks}{\ensuremath{\mathcal{S}}}
\providecommand{\graphsdk}{\ensuremath{S}}

\begin{document}


\begin{frontmatter}


\paperid{123} 


\title{Automated Planning for Optimal Data Pipeline Instantiation}


\author[E,A]{\fnms{Leonardo}~\snm{Rosa Amado}\thanks{Corresponding Author. Email: leonardo.amado@abdn.ac.uk.}}
\author[A,B]{\fnms{Adriano}~\snm{Vogel}\footnote{Work done while affiliated with the Pontifical Catholic University of Rio Grande do Sul (PUCRS).}}
\author[A]{\fnms{Dalvan}~\snm{Griebler}}
\author[C]{\fnms{Gabriel}~\snm{Paludo Licks}}
\author[D]{\fnms{Eric}~\snm{Simon}} 
\author[E,A]{\fnms{Felipe}~\snm{Meneguzzi}}

\address[A]{Pontifical Catholic University of Rio Grande do Sul, Brazil}
\address[B]{Johannes Kepler University Linz, Austria}
\address[C]{Sapienza University of Rome, Italy}
\address[D]{SAP Labs, France}
\address[E]{University of Aberdeen, Scotland}


\begin{abstract}
Data pipeline frameworks provide abstractions for implementing sequences of data-intensive transformation operators, automating the deployment and execution of such transformations in a cluster. 
Deploying a data pipeline, however, requires computing resources to be allocated in a data center, ideally minimizing the overhead for communicating data and executing operators in the pipeline while considering each operator's execution requirements. 
In this paper, we model the problem of optimal data pipeline deployment as planning with action costs, where we propose heuristics aiming to minimize total execution time. 
Experimental results indicate that the heuristics can outperform the baseline deployment and that a heuristic based on connections outperforms other strategies.
\end{abstract}

\end{frontmatter}


\section{Introduction}
\label{sec:intro}


A recent report from the European Commission forecasts an increase to 8.4 million data professionals in the EU27 by 2025, with 1.8 million positions to be added in the period between 2020 and 2025, leading to a shortage of approximately 484,000 positions~\cite{Cattaneo2020datamarket}. 
The shortage of highly trained human resources calls for data processing platforms accessible to a broader category of users (the so-called citizen data scientists) having strong data expertise, moderate statistics training, and low expertise in programming and distributed computing~\cite{burns2018,pi2019}. 
Several recent technology trends acknowledge this situation through the development of 
data pipeline frameworks~\cite{paleyes2021, Mahapatra2020, noflo, nussnacker, flowpipe, nodered} relying on a flow-based programming model \cite{Morrison1994}.

In these frameworks, the goal is to enable non-technical experts (aka citizen data scientists) to quickly build, debug, and maintain high-quality data pipelines at scale within a cloud cluster by providing high-level abstractions in their programming model. 
Since users design data pipelines, and are not assumed to comprehend the underlying engine that deploys and executes the pipeline, the resulting data pipelines (graphs) derive a non-optimal allocation of resources to execute it. 
Thus, data pipeline engines must provide optimization techniques to assure an optimal usage of computing resources in a distributed system and minimize data-intensive communication costs during graph execution. 

The problem of optimizing a pipeline execution is similar to the task of constraint optimization~\cite{hirzel2014}, which has  been a focus of research on automated planning. Indeed, research in that area has focused on developing algorithms to compute optimized solutions for optimization problems with constraints, leading to fast and reliable algorithms~\cite{Haslum:2019gc}. 
%
%
With such a research background, we develop a model for automated planning algorithms that use heuristic search and state-dependent action costs to solve the problem of optimally allocating computation tasks to the worker nodes of a distributed system for executing a data pipeline. 
We evaluate our optimization methods using the pipeline engine of the SAP Data Intelligence  commercial product \footnote{\url{https://www.sap.com/products/technology-platform/data-intelligence.html}}, called VFlow \footnote{\url{https://api.sap.com/api/vflow/overview}}, which adopts a flow-based programming model, and whose execution model is based on a distributed Kubernetes container orchestration platform.

This work provides two main technical contributions: I) We model several optimization methods that determine how the operators of a data pipeline should be deployed on the worker nodes of Kubernetes clusters; II) We perform an experimental analysis of the proposed approaches with a real-world industry data pipeline processing system, report our findings and indicate directions for future performance improvement. 

We organize this article as follows. 
Section~\ref{sec:background} provides an overview of automated planning and data pipelines.
In Section~\ref{sec:problem}, we model the elements and constraints of deploying data pipelines and computing plans. Then, in Section~\ref{sec:solution} we describe the approaches to pipeline grouping for reducing the total execution time. 
Section~\ref{sec:methodology} explains our experimental methodology that compares the results of the heuristic approaches against solutions from a random baseline approach on various workloads. 
Section~\ref{sec:results} shows our empirical performance results for the different approaches using pipelines with a sequential or parallel topology in which we vary the number of special operators and the computational load. 
Finally, Section~\ref{sec:conclusion} summarizes our key contributions and points toward future work.


\section{Background}
\label{sec:background}


This section introduces the formal background upon which our application relies. 
We start with a formalization of Automated Planning and the elements of problems in this formalism in Section~\ref{sec:planning}. 
Section~\ref{sec:pipelines} then formalize data pipelines and how their definition relates to deployment in an underlying high performance infrastructure. 
This allows us to formalize the pipeline optimization problem of Section~\ref{sec:problem}.

\subsection{Automated Planning}
\label{sec:planning}

We leverage automated planning techniques to drive the optimization. 
Automated planning is a sub-area of Artificial Intelligence concerned with finding sequences or policies of actions (\idest, plans) able to transition an agent from a given initial state to a desirable goal state~\cite[Ch. 1]{Geffner:2013hx,chowdhary2020}. 
A classical planning problem can be described as one of graph search, in which the nodes in the graph represent states, edges represent transitions between states that are caused by applying an action at a state. 
The solution to a planning problem is a sequence of actions (edges) that forms a path to traverse this graph from a unique initial state to any one of the goal states. 
Domain experts seldom define such graphs explicitly, but rather use a logic-based description language (which we define later) to induce the actual search graph. 
This allows graphs with huge state-spaces to be defined compactly. 
In what follows we adopt the planning terminology from Ghallab~\etal~\cite{AutomatedPlanning_Book2016} to represent states and actions in planning domain problems.

The most fundamental part of a classical planning task is the planning domain. 
A planning domain is a formal description of the dynamics of an environment in which an agent acts. 
Definition~\ref{def:planningdomain} formalizes a planning domain.
\begin{definition}[\textbf{Planning Domain}]\label{def:planningdomain}
A planning domain  $\planningdomain$ consists of a pair $\tuple{\fluents,\actions}$, which specifies the knowledge of the domain, and comprises: a finite set of facts $\fluents$, \idest, a set of ground instantiated predicates, defining the environment state properties; and
a finite set of actions $\actions$, which is technically a set of ground instantiated operators, representing the actions that can be performed in the environment.

\end{definition}
States $s \subseteq \fluents$ consist of facts from a planning domain indicating properties that are true at any moment in time and follow the closed world assumption so that any fact not included in a state is false.
Conditions or formulas in our formalism comprise positive and negated facts ($\fluent$, $\neg\fluent$) representing an implicit conjunctive formula indicating what must be true (alternatively, false) in a state. 
The positive part of a condition $\condition$ ($\positive(\condition)$) comprises the positive facts, and the negated part of a condition $\negative(\condition)$ comprises the negated facts. 
We say a state $s$ supports a condition $\condition$, $s \models c$ (alternatively, $c$ is valid in $s$), if and only if all positive facts are present in $s$, and all negated facts are absent in $s$, i.e., $s \models c$ if and only if $(s \cup \positive(\condition) = s) \land (s \cap \negative(\condition) = \emptyset)$. 
An action $\action \in \actions$ is represented by a tuple $o = \tuple{\pre(\action),\eff(\action),\cost(\action)}$ containing the preconditions $\pre(\action)$, the effects $\eff(\action)$, and a non-negative cost $\cost(\action)$, indicating possible transitions between states. 

The transition of a state $s$ into a new state $s'$ using an action $\action$ is represented as the transition function $s' = \transition(s,\action)$. 
The transition is valid iff $s \models \pre(\action)$, and $s' = \left(s \cup \positive(\eff(\action))\right) - \negative(\eff(\action))$. 

A planning task (or planning instance) is the formal description of a task to be solved in a given planning domain. 
Planning tasks comprise a planning domain $\planningdomain$ and a planning problem. 
A planning problem describes the finite set of objects of the environment, the initial state from which the planning problem starts, and the goal state which an agent desires to achieve. 
\begin{definition} [\textbf{Planning Instance}]\label{def:planningtask}
A planning task is a tuple $\planningtask = \tuple{\planningdomain,\initialstate,\goalcondition}$, in which $\initialstate$ is an initial state, $\goalcondition$ is a goal condition
, and $\planningdomain$ is a planning domain. 
\end{definition}

The solution for a planning task, which we call a \textit{plan}, is a sequence of actions that an agent can perform in a planning domain to achieve the goal state $\goalcondition$ from the initial state $\initialstate$. 
Definition~\ref{def:plan} formalizes the notion of a plan and an optimal plan.

\begin{definition} [\textbf{Plan}]\label{def:plan}
A plan $\plan$ for a planning task $\planningtask$ is a sequence of actions $\pi = \tuple{\action_1, \dots, \action_n}$ that induces a sequence of states $\tuple{s_0, s_1, \dots, s_n}$ such that $\initialstate = s_0 \models \pre(\action_1)$, $s_n \models \goalcondition$ and that every state $s_i \in \plan$ is such that $s_{i-1} \models \pre(\action_i)$ and $s_{i} = \transition(s_{i-1},a_{i})$. 
The cost of a plan is the sum of the cost all of its actions such that $\cost(\plan) = \displaystyle\sum_{i=1}^{n} \cost(\action_i)$. 
We say a plan is optimal, denoted $\optimalplan$, if its cost is minimal, that is, if no other plan $\plan$ has $\cost(\plan) < \cost(\optimalplan)$. 
\end{definition}

Note that Definition~\ref{def:plan} allows for multiple optimal plans, which may have the same cost, and are thus equivalently optimal. 
Modern planners use the Planning Domain Definition Language (PDDL) as a standardized domain and problem representation medium~\cite{Fox2003}. 
PDDL encodes a number of expressive planning formalisms, such as STRIPS-style~\cite{Fikes1971} planning tasks. 
For simplicity, most planning formalizations assume that every action in $\actions$ has cost 1, so that the optimal plan is the plan with the smallest number of actions. 
Here, we define actions with varying costs that depend on the state of the environment, where the environment considered in this work relates to the configurations of a platform to run data pipelines. 

\subsection{Designing and deploying data pipelines}
\label{sec:pipelines}

In the flow-based programming model~\cite{Morrison1994}, a data pipeline corresponds to a type of application that processes large amounts of data and produces results promptly. We refer to processing as the capacity of data pipelines to ingest, transform, and output data with high throughput and low latency~\cite{hueske2019}.
At the design level, the data pipelines considered in this article are  graphs in which nodes are operators and edges represent the flow of data, following the flow-based programming paradigm \cite{Morrison1994}. 
Operators\footnote{Note that we explicitly make a distinction between \textit{operators} (i.e., elements of the data pipeline formalism) and \textit{actions} (i.e., state transformation functions in the planning formalism)} are responsible for performing operations with data.  
Operators can be written using different programming languages using a Software Development Kit (SDK) provided by the data pipeline framework, formalized as Definition~\ref{def:sdk}. 

\begin{definition}[SDK]\label{def:sdk}
    Let $\graphsdks$ be the set of all SDKs made available by the pipeline engine.
	An SDK $\graphsdk \in \graphsdks$ provides a programming environment and runtime specific to a programming language, with tools such as an API for the user to program operators and interact with the pipeline framework.
	Thus, the data pipeline framework can provide SDKs for developing operators in Java, Python, Go, etc. 
\end{definition}

Implementations of operators can differ in requirements that need to be satisfied by the runtime system in order to be executable, and \textit{tags} associated with an operator encode these requirements. 
Definition~\ref{def:tag} formalizes the notion of a tag. 

\begin{definition}[Tag]\label{def:tag}
	Let $\graphtags$ be the set of all possible tags defined by the user.
	A tag $\graphtag \in \graphtags$ denotes a requirement that has to be satisfied for an operator to be executable, i.e. a specific library or package (possibly a specific version of each).
\end{definition}

Definition~\ref{def:operator} brings together Definitions~\ref{def:sdk} and~\ref{def:tag} into the formal operator within a pipeline.

\begin{definition}[Operator]\label{def:operator}
	A pipeline operator $\graphoperator=\tuple{\graphoperatorid, \graphsdk, \operatortags}$ comprises an identification $\graphoperatorid$, the SDK $\graphsdk \in \graphsdks$ it is developed with, and a set of tags $\operatortags \subseteq \graphtags$ that it requires in order to execute. 
\end{definition}

The runtime environment in which an operator must be executed is set by creating a Docker image (later called OS image)~\cite{turnbull2014}, which is is a lightweight way to package code, run-time and all dependencies, and by associating tags to the OS image that reflect the characteristics of the environment. 
Definition~\ref{def:image} formalizes the notion of an image in terms of the tag requirements from Definition~\ref{def:tag}.

\begin{definition}[Image]\label{def:image}
	An image $\graphimage \subseteq \graphtags$ is a set of tags that the image specification supports, representing a possible grouping for operators' tags satisfied by the image.
\end{definition}

A designer of a data pipeline can then tag an operator to indicate the runtime environment required for its execution. 
By default, when deploying a data pipeline for execution, the execution framework searches for an OS image whose tags match all the required tags of the operators in the data pipeline. 
In this case, all operators become part of a default ``group''. 
Operators in a group are deployed as a container running in a Kubernetes \textit{Pod}~\cite{bernstein2014containers}: they share the libraries and packages available in the environment defined by the containerized OS image. 
Definition~\ref{def:group} formalizes such matching of tags to images as a group.

\begin{definition}[Group]\label{def:group}
    Let $\graphgroups$ be the set of groups in a data pipeline and $\graphoperators$ be the set of operators in a data pipeline.
    A group $\graphgroup \in \graphgroups$ then is a tuple $\tuple{\groupoperators, \graphimage}$, where $\groupoperators \subseteq \graphoperators$ is the set of operators in the group and $\graphimage \in \graphimages$ is the image associated to the group.
    Since all operators have to be grouped, explicitly or implicitly, the intersection between operators in each group has to be empty, i.e. $\groupoperators{1} \cap \groupoperators{2} = \emptyset$ for all $\graphgroup_{i} \in \graphgroups$. 
\end{definition}

However, when the execution framework cannot find a single matching OS image, the user must manually group operators according to their required tags until the remaining non grouped operators can be matched by a single OS image. 
Alternatively, in order to have control over which requirements a user needs for specific tasks, users can manually group operators and create an OS image specific to the grouped operators. 
Finally, we bring together the notion of an undeployed Data Pipeline incorporating the elements from Definitions~\ref{def:sdk}--\ref{def:operator} in Definition~\ref{def:dataPipeline}.

\begin{definition}[Data Pipeline]\label{def:dataPipeline}
	Let $\graphedge = \tuple{\graphoperator_{i},\graphoperator_{j}}$ be an edge tuple representing the directed connection between two operators $\graphoperator_{i}$ and $\graphoperator_{j}$. 
	A data pipeline then is a tuple $\datapipeline = \tuple{\graphgroups, \graphedges}$, containing the specified groups and edges between all operators.
\end{definition}

\begin{figure}[t]
    \centering
    \includegraphics[width=0.4\textwidth]{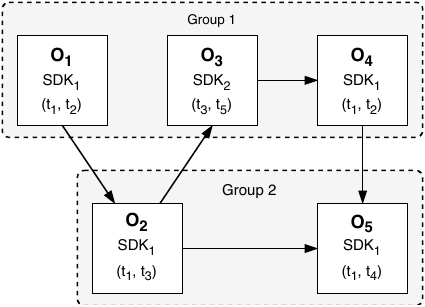}
    \vspace{0.2cm}
    \caption{Data pipeline illustration at design level}
    \vspace{0.85cm}
    \label{fig:designLevelGraph}
\end{figure}

Figure~\ref{fig:designLevelGraph} exemplifies a data pipeline with operators grouped using different requirements, specified by tags. 
Here, the data pipeline is executable if the runtime framework has at least two images $\graphimage_{1}$ and $\graphimage_{2}$, such that $\tuple{t_{1}, t_{2}, t_{3}, t_{5}} \subseteq \graphimage_{1}$ and $\tuple{t_{1}, t_{3}, t_{4}} \subseteq \graphimage_{2}$.


\section{The pipeline optimization problem}
\label{sec:problem}

At design level, data pipelines are graphs in which nodes correspond to operators and edges represent the flow of data. 
There may be multiple concrete implementations of data pipelines\footnote{We refer to data pipelines as graphs and use the term interchangeably.} and many open source systems do, including Flink~\cite{Carbone:2015ws}, Storm and others. 
In this work, we focus specifically on the implementation provided in the VFlow engine where operators perform computing operations with data. 
Operators in VFlow can be written using different programming languages using SDKs provided by the data pipeline framework. 

Given a data pipeline $\datapipeline$ from Definition \ref{def:dataPipeline}, with a set of connected operators, we must find the optimal grouping configuration, in terms of total execution time, which satisfies the constraints imposed by the operator tags $\operatortags$. 
It is therefore an optimization problem under constraints, where the constraints are the tags of operators, which must match the tags of an OS image to belong to the same group. 



The runtime environment in which operators execute can be set by creating groups of operators. 
These runtime environments are executing on container images (e.g., Docker images) where each group ideally has the same image associated to it. 
This allows the operators of the same group share resources, libraries, and packages available in that environment.
In order to have control over which requirements a user needs for specific tasks, 
Users can manually group operators and create an image specific to the grouped operators. 
In the default group, VFlow associates available images and stock images that are tailored to the basic requirements of each SDK.
In case operators are not explicitly grouped by users they belong to an implicit group such that the pipeline framework automatically associates an image to it using the available images and stock images that are tailored to basic requirements of each SDK. 


\section{Automating operator grouping}
\label{sec:solution}


In this section, we detail four optimization methods: two greedy approaches and one approach based on deriving specific costs. 
As a baseline, we define a random approach that optimizes the pipeline by selecting random images and adding random operators. 
We use PDDL to create descriptions of state-space search problems involving a variety of domains independently of the underlying AI search engine, enabling us to solve the problem of Definition~\ref{def:planningtask}. 
Such flexibility, coupled with the number of efficient planning software available~\cite{Scala:2016wy,enhsp}, allows us to prototype a number of optimization strategies for grouping in order to achieve desirable deployment properties. 

\begin{figure}[t]
    \centering
    \includegraphics[width=0.46\textwidth]{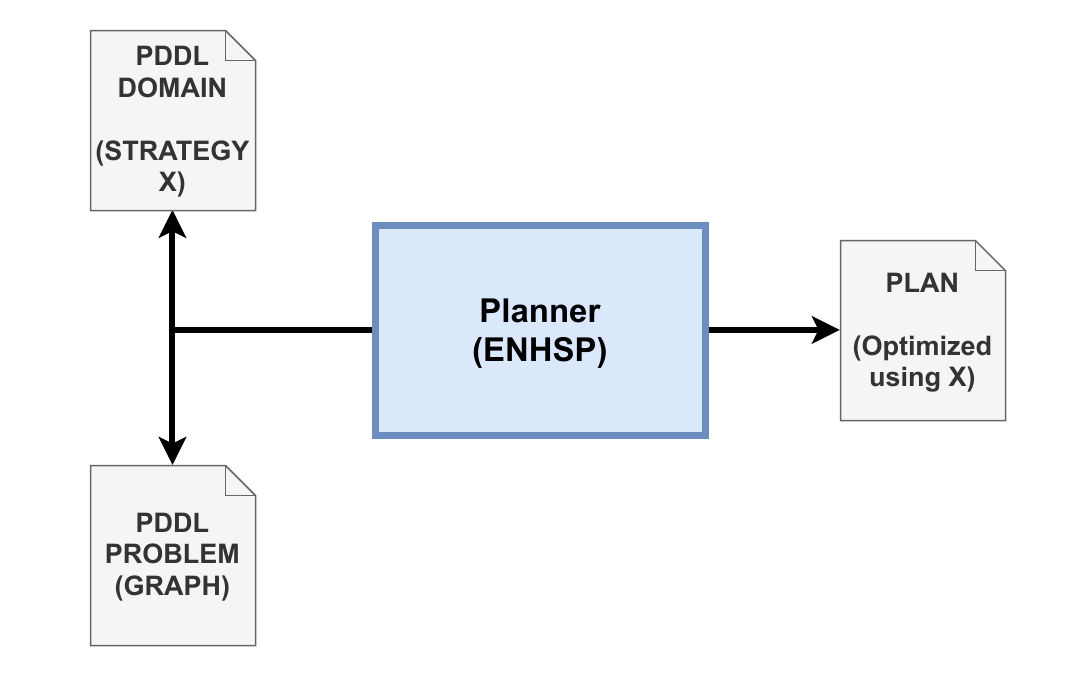}
    \vspace{0.1cm}
    \caption{Planner workflow.}
    \vspace{0.755cm}
    \label{fig:planner_workflow}
\end{figure}

In order to generate an optimized graph, we provide two such specifications to a PDDL planner such as ENHSP~\cite{Haslum:2019gc}, and the resulting plan can then be translated into an optimized graph. We illustrate this workflow in Figure~\ref{fig:planner_workflow}. 

\begin{figure}[b]
    \centering
    \begin{subfigure}[t]{0.4\textwidth}
        \centering
        \includegraphics[width=\textwidth]{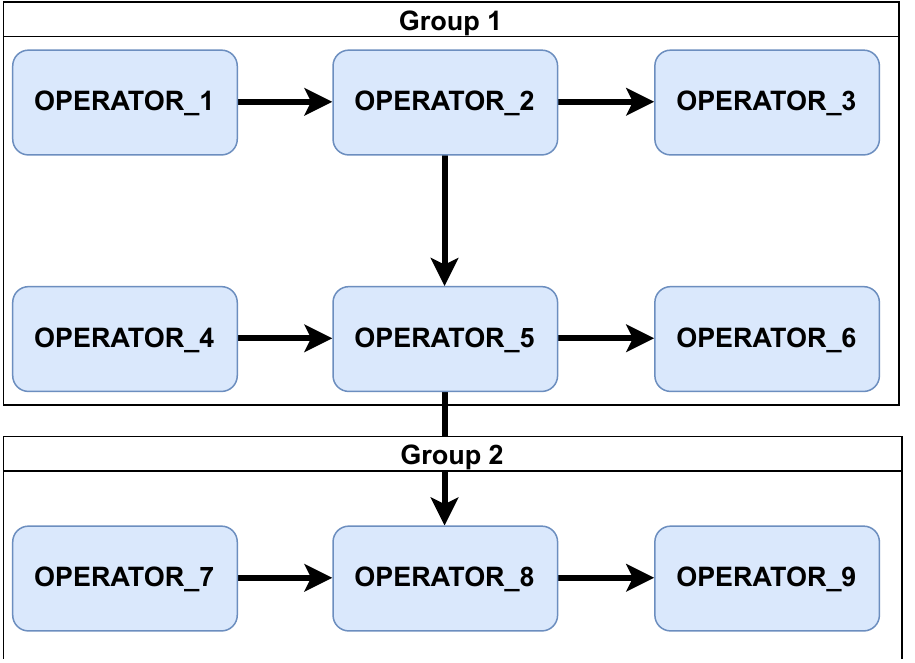}
        \caption{Connection Heuristic.}
        \vspace{0.6cm}
        \label{fig:conn_heu}
    \end{subfigure}
    \begin{subfigure}[t]{0.4\textwidth}
        \centering
        \includegraphics[width=\textwidth]{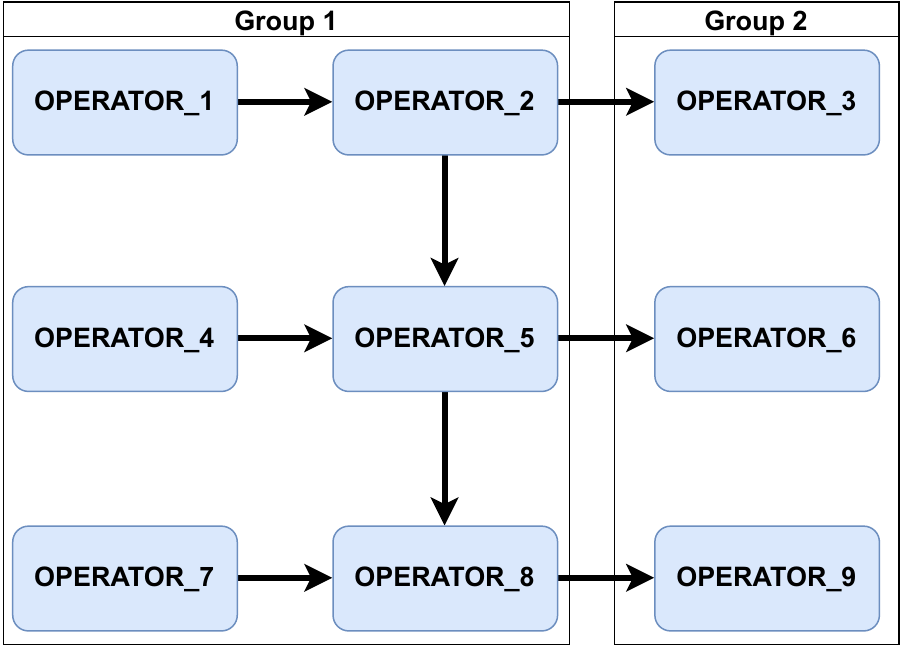}
        \caption{Node Heuristic.}
        \label{fig:node_heu}
    \end{subfigure}
    \vspace{0.5cm}
    \caption{Proposed heuristics.}
    \vspace{0.5cm}
    \label{fig:heuristics}
\end{figure}

\subsection{Connection heuristic}
\label{sec:connection_heuristic}

The \emph{connection} greedy approach prioritizes the creation of groups with a minimum number of intergroup communication links.  
To encode this approach using PDDL, we use different weights for intragroup and intergroup communication, respectively $5$ and $20$. 
We assign a medium cost for instantiating groups, since creating many groups will lead to more intergroup connections. 
Figure~\ref{fig:conn_heu} exemplifies a solution using this approach, where the number of intergroup communication links is only one and two operator groups are created.         
  
\subsection{Node heuristic}
\label{sec:node_heuristic}

The \emph{node} approach prioritizes the creation of groups with as many nodes as possible, greedily searching for groups with a larger number of operators. 
To encode this approach using PDDL, we use the same weights for intragroup and intergroup communication, respectively $5$.  
We assign a very high cost for instantiating groups, which forces the planner to search for configurations with a very small number of groups. 
Figure~\ref{fig:node_heu} illustrates an example of a solution using this heuristic. 
As we can see, this solution ignores the number of intergroup connections, leading to two groups with many intergroup connections.
             
\begin{figure}[t]
    \centering
    \begin{subfigure}[b]{0.47\textwidth}
        \centering
        \includegraphics[width=\textwidth]{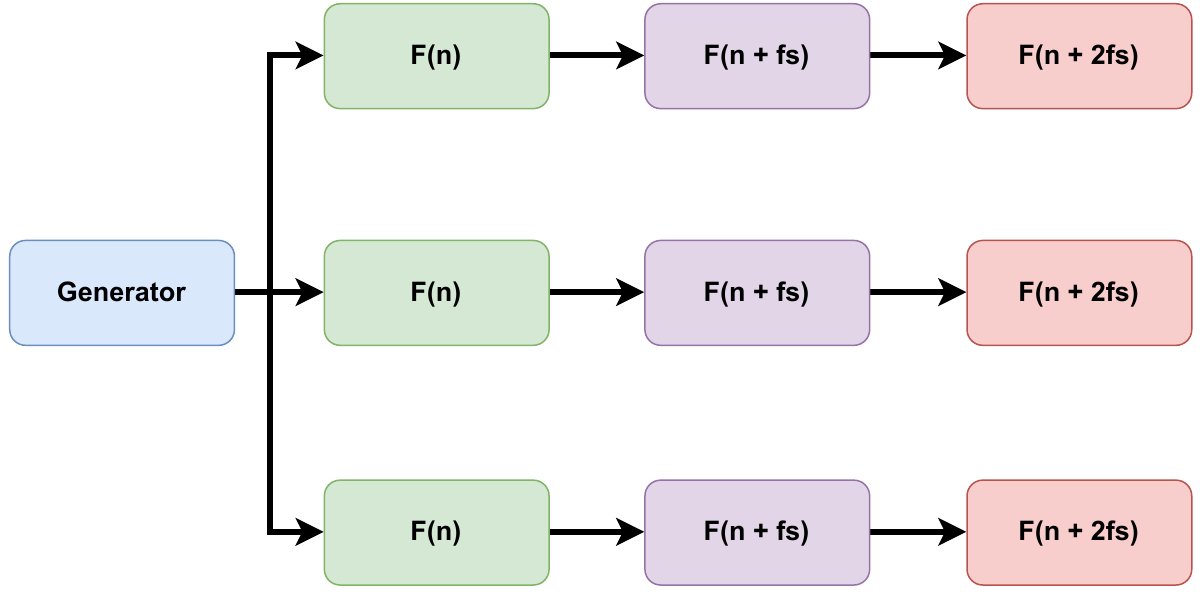}
        \caption{Fibonacci step parameter in a parallel graph.}
        \vspace{0.7cm}
        \label{fig:fibo_step}
    \end{subfigure}
    \begin{subfigure}[b]{0.47\textwidth}
        \centering
        \includegraphics[width=\textwidth]{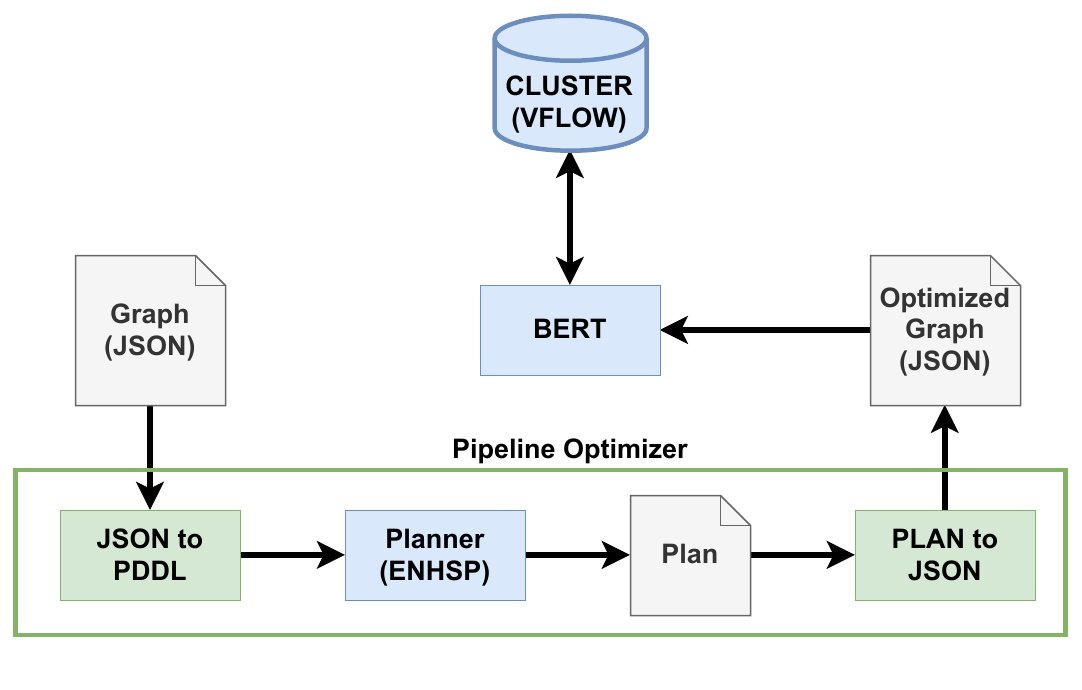}
        \vspace{-0.6cm}
        \caption{Pipeline optimizer framework.}
        \label{fig:architecture}
    \end{subfigure}
    \vspace{0.6cm}
    \caption{Fibonacci and pipeline optimizer.}
    \vspace{0.8cm}
    \label{fig:evaluation} 
\end{figure}

\subsection{Random baseline}
The random baseline approach groups operators by randomly selecting an image from the list of available images and then grouping all operators satisfied by the selected image, if there are any.
This algorithm randomly selects images and groups operators until no operators are left to be grouped. 
This process of randomly selecting groups works as follows. 
First, we randomly select an available image capable of executing operators based on the available tags of this image. 
Second, we select all $n$ operators that can run using this image. 
Third, if the number of operators is greater than 0, we randomly compute a number from the range $[1,n]$, which defines how many of these possible operators our new group will contain. 
Finally, we generate a new group with the selected operators and repeat this process until all operators are assigned to a group.





\section{Evaluation methodology}
\label{sec:methodology}

The evaluation of our solution requires that we design different examples of data pipelines and generate a synthetic workload that incur substantive computing.
We instantiate data pipelines following the two most common types of patterns that users design: sequential and parallel executions.
As a workload choice, we choose the computation of Fibonacci sequences, which allows us to parameterize the amount of work that needs to be computed by an operator by defining the size of the sequence.
We detail these in the following sections.

\subsection{Workload configuration}
\label{sec:setup}

\begin{figure}[t]
    \centering
    \begin{subfigure}[c]{0.36\textwidth}
        \centering
        \includegraphics[width=\textwidth]{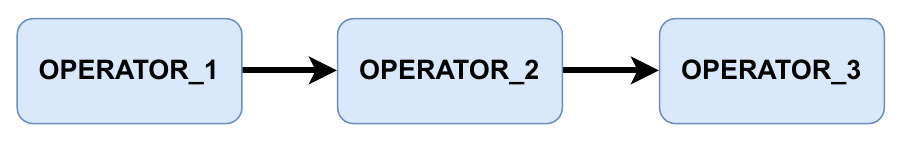}
        \caption{Single line}
        \vspace{0.7cm}
        \label{fig:singleLineGraph}
    \end{subfigure}
    \begin{subfigure}[c]{0.47\textwidth}
        \centering
        \includegraphics[width=\textwidth]{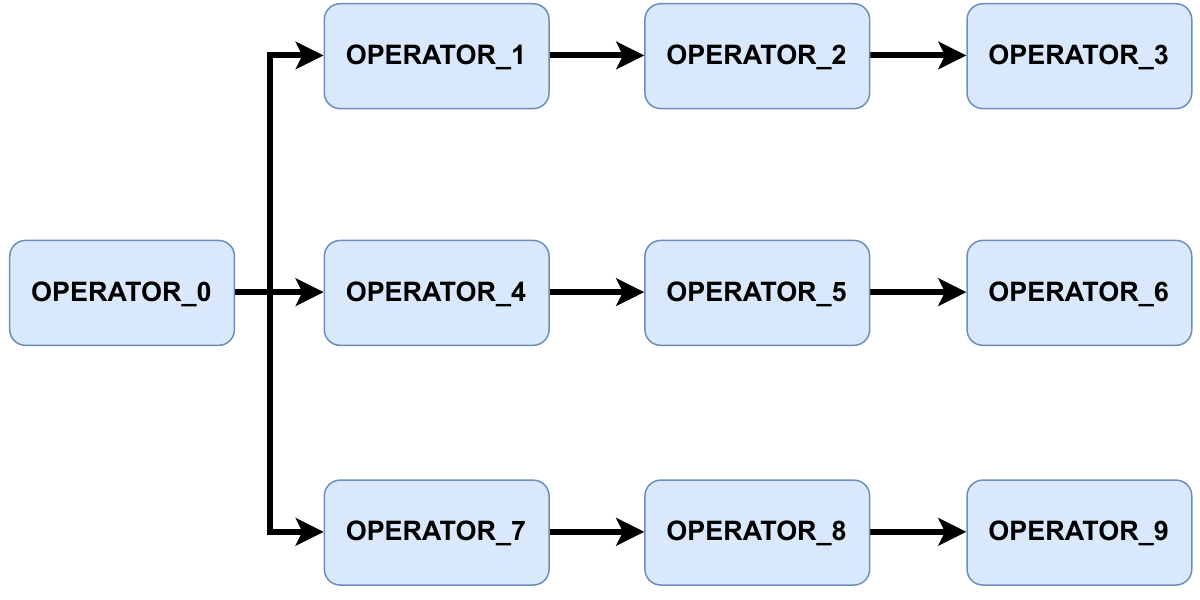}
        \caption{Parallel lines}
        \label{fig:parallelLinesGraph}
    \end{subfigure}
    \vspace{0.6cm}
    \caption{Graph topologies used in evaluation.}
    \vspace{0.8cm}
    \label{fig:graphTopologies}
\end{figure}

In order to evaluate the grouping methods, we use two patterns of synthetic data pipelines built from the same operator which computes the $n^{th}$ number of the Fibonacci sequence and sends it to the next operator.  
This operator allows us to control the processing time of a data pipeline. 
The two patterns of data pipelines, illustrated in Figure~\ref{fig:graphTopologies}, are as follows:

\begin{enumerate}
	\item a topology representing sequential computation steps;
	\item a topology representing independent and concurrent paths of sequential computation steps. 
\end{enumerate}

We define two parameters to generate pipeline optimization problems with varying degrees of complexity. 
These parameters are: \textbf{number of special tags} (later called \emph{special\_ops}) and \textbf{Fibonacci step}. 



The \textbf{special tag} represents operators with specific library requirements beyond what is included in the standard images associated with operators in each sub-engine. Those necessitate specific images to be run in VFlow. By default, all operators have default \textit{"golang"} tags, and all images have the default tag. In addition to the default tag \textit{"golang"}, we create three distinct special tags (\textit{spt-1}, \textit{spt-2}, and \textit{spt-3}), where each special tag belongs to a single specific image. When the parameter "special\_ops" has the value $n$, then $n$ randomly selected operators will receive a special tag. If $n \leq 3$, then all the operators with special tags will have distinct tags between (\textit{spt-1}, \textit{spt-2}, \textit{spt-3}). Otherwise, two operators can have the same special tag. For example, if $n = 4$, two operators will have the special tag \textit{spt-1}. 

The goal of the \textbf{special tag} is to increase the complexity of the optimization problem. 
By default, all generated pipelines already contain one special tag to ensure the problem cannot be trivially solved by including all operators in a single group. 
As the number of special tags increases, the problem of grouping the operator becomes more difficult.

\begin{figure}[h!]
    \centering
    \begin{subfigure}[b]{0.49\textwidth}
        \centering
        \includegraphics[width=\textwidth]{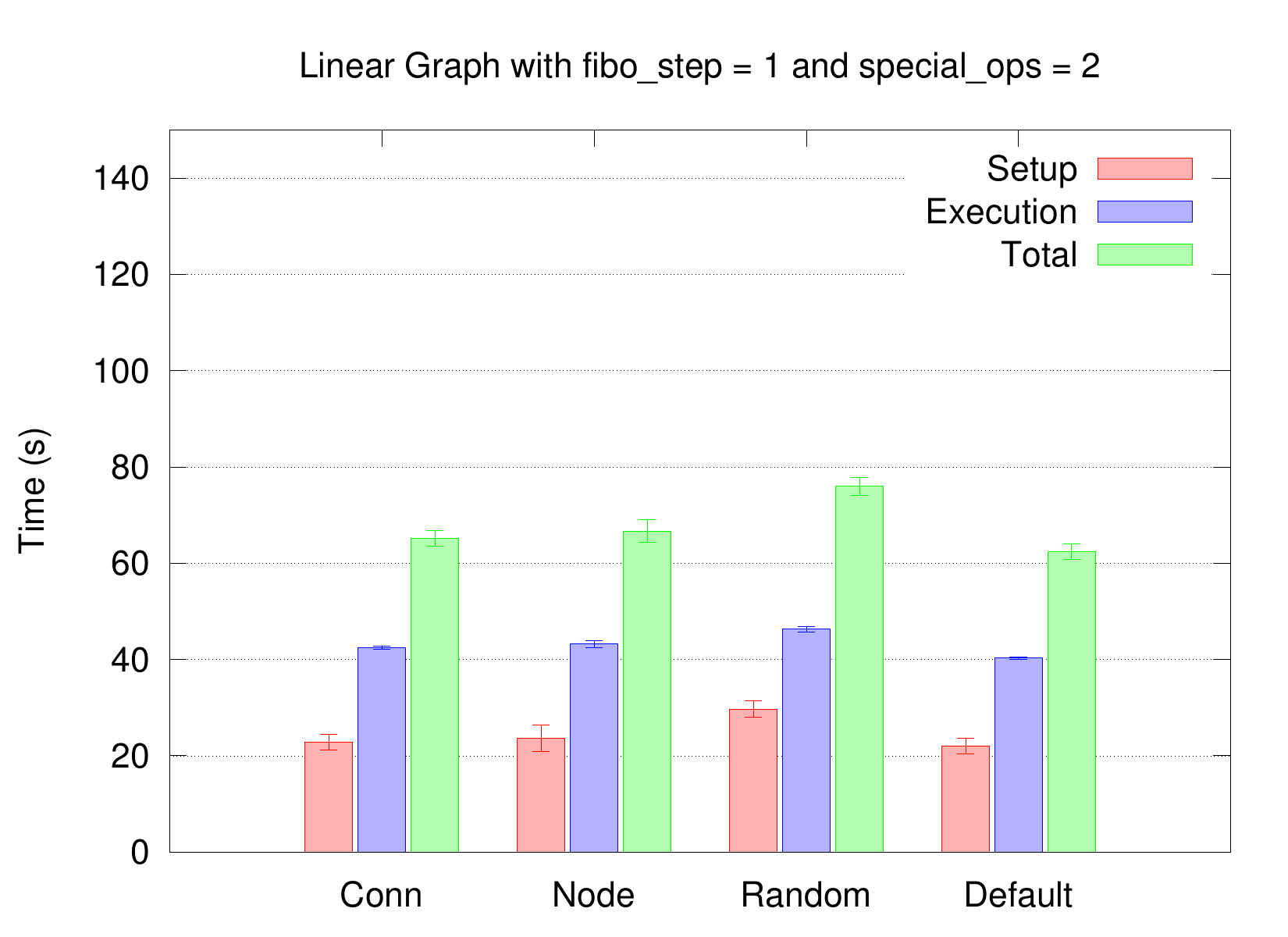}
        \caption{Line topology. Step = 1, sp-ops = 2.}
        \vspace{0.4cm}
        \label{fig:line_st1so1}
    \end{subfigure}
    \begin{subfigure}[b]{0.49\textwidth}
        \centering
        \includegraphics[width=\textwidth]{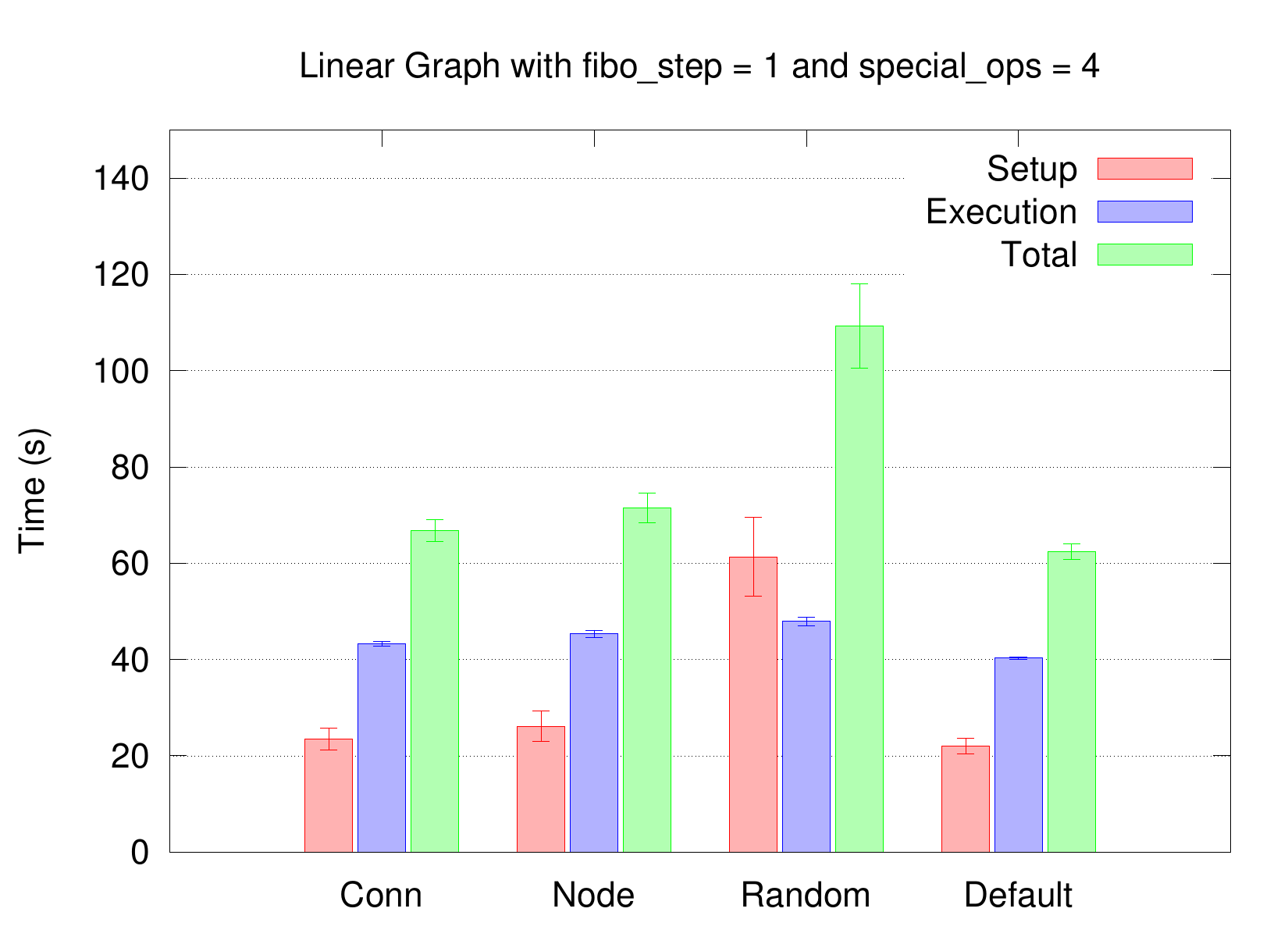}
        \caption{Line topology. Step = 1, sp-ops = 4.}
        \vspace{0.4cm}
        \label{fig:line_st1so3}
    \end{subfigure}
    \begin{subfigure}[b]{0.49\textwidth}
        \centering
        \includegraphics[width=\textwidth]{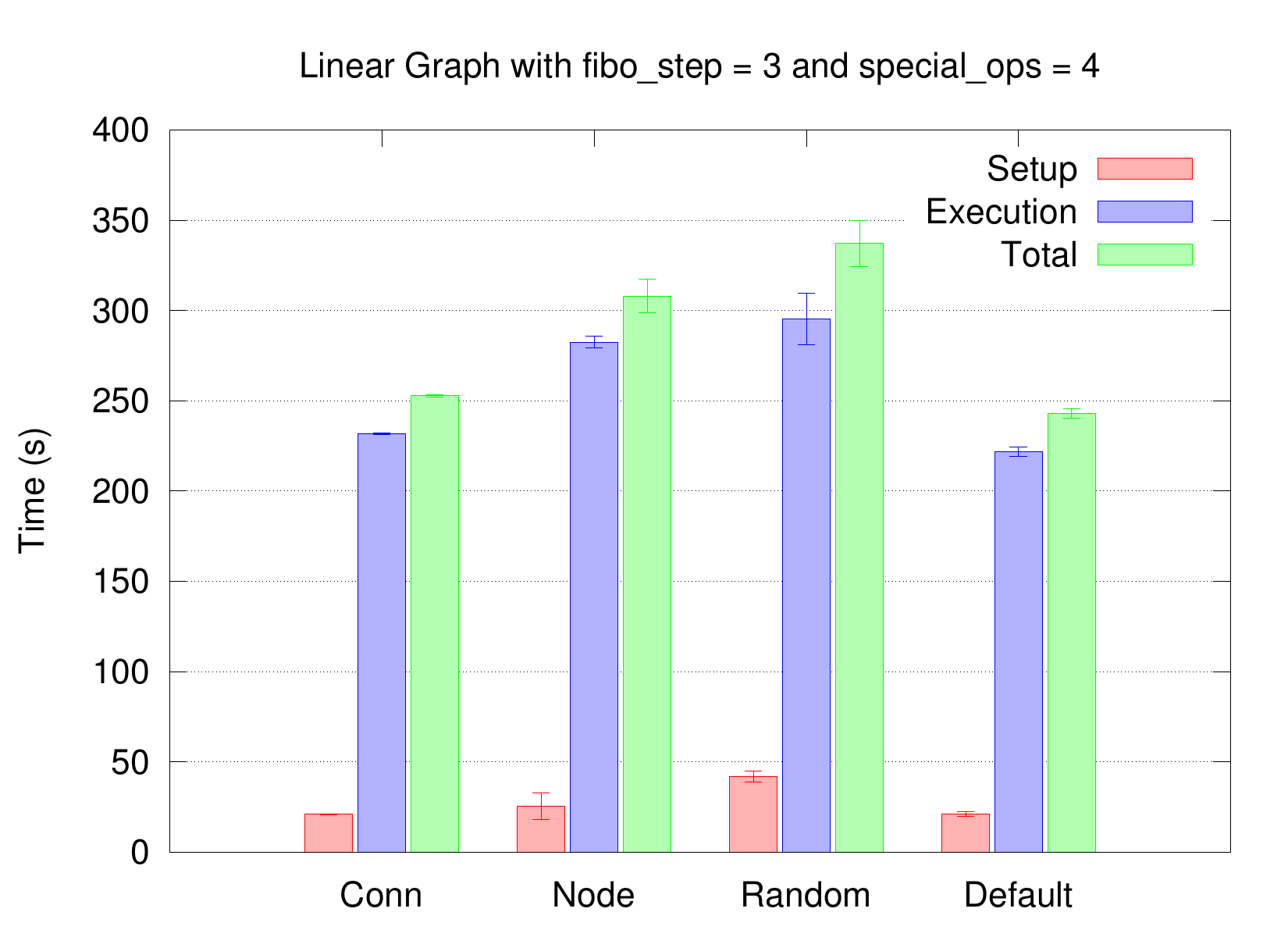}
        \caption{Line topology. Step = 3, sp-ops = 4.}
        \vspace{0.6cm}
        \label{fig:line_st3so3}
    \end{subfigure}
    \caption{Experiment results for Line topology.}
    \vspace{0.4cm}
    \label{fig:exp0}
\end{figure}

\begin{figure}[h!]
    \centering
    \begin{subfigure}[b]{0.49\textwidth}
        \centering
        \includegraphics[width=\textwidth]{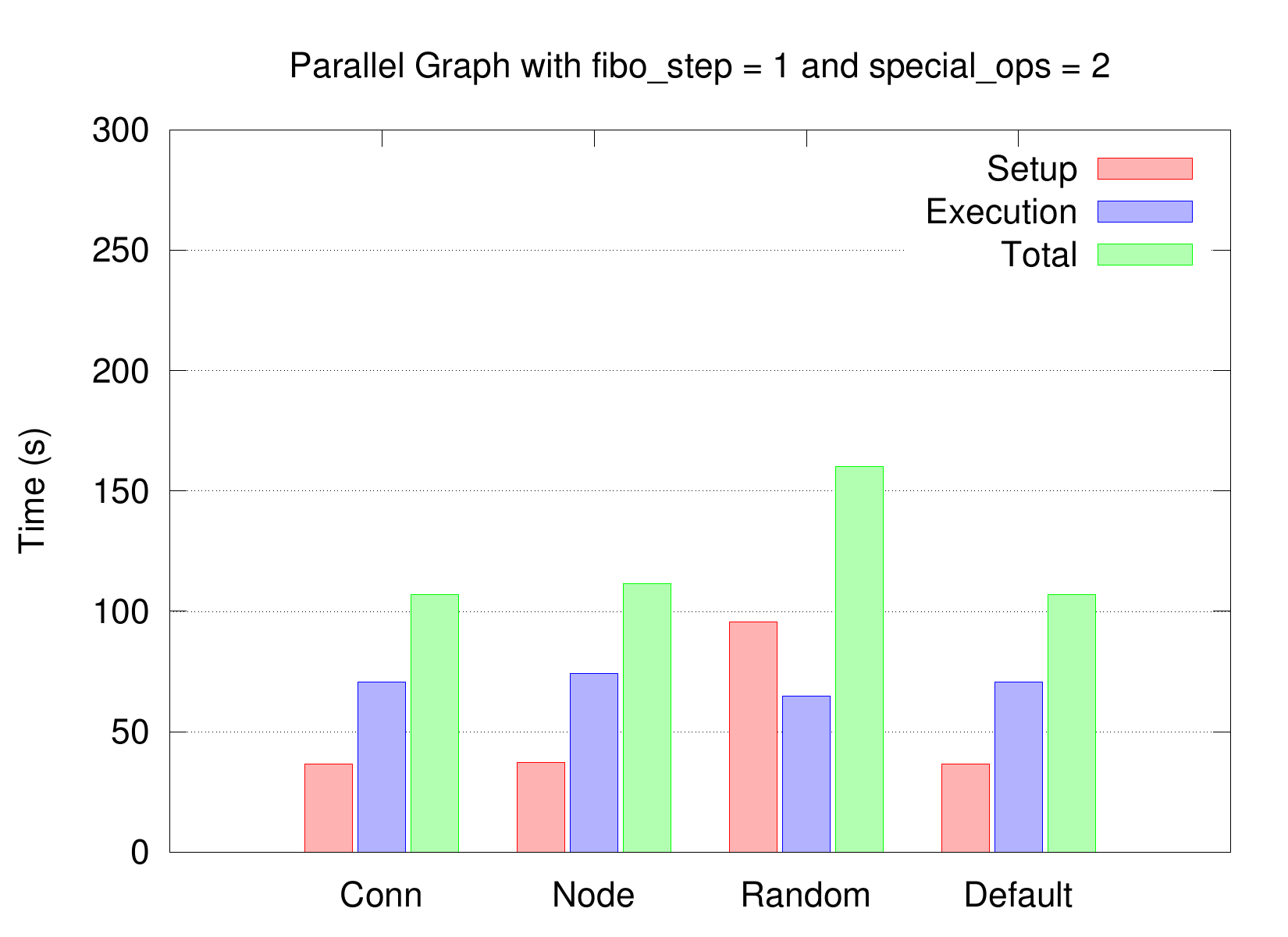}
        \caption{Parallel. Step = 1, sp-ops = 2.}
        \vspace{0.4cm}
       \label{fig:para_st1so1}
    \end{subfigure}
    \begin{subfigure}[b]{0.49\textwidth}
        \centering
        \includegraphics[width=\textwidth]{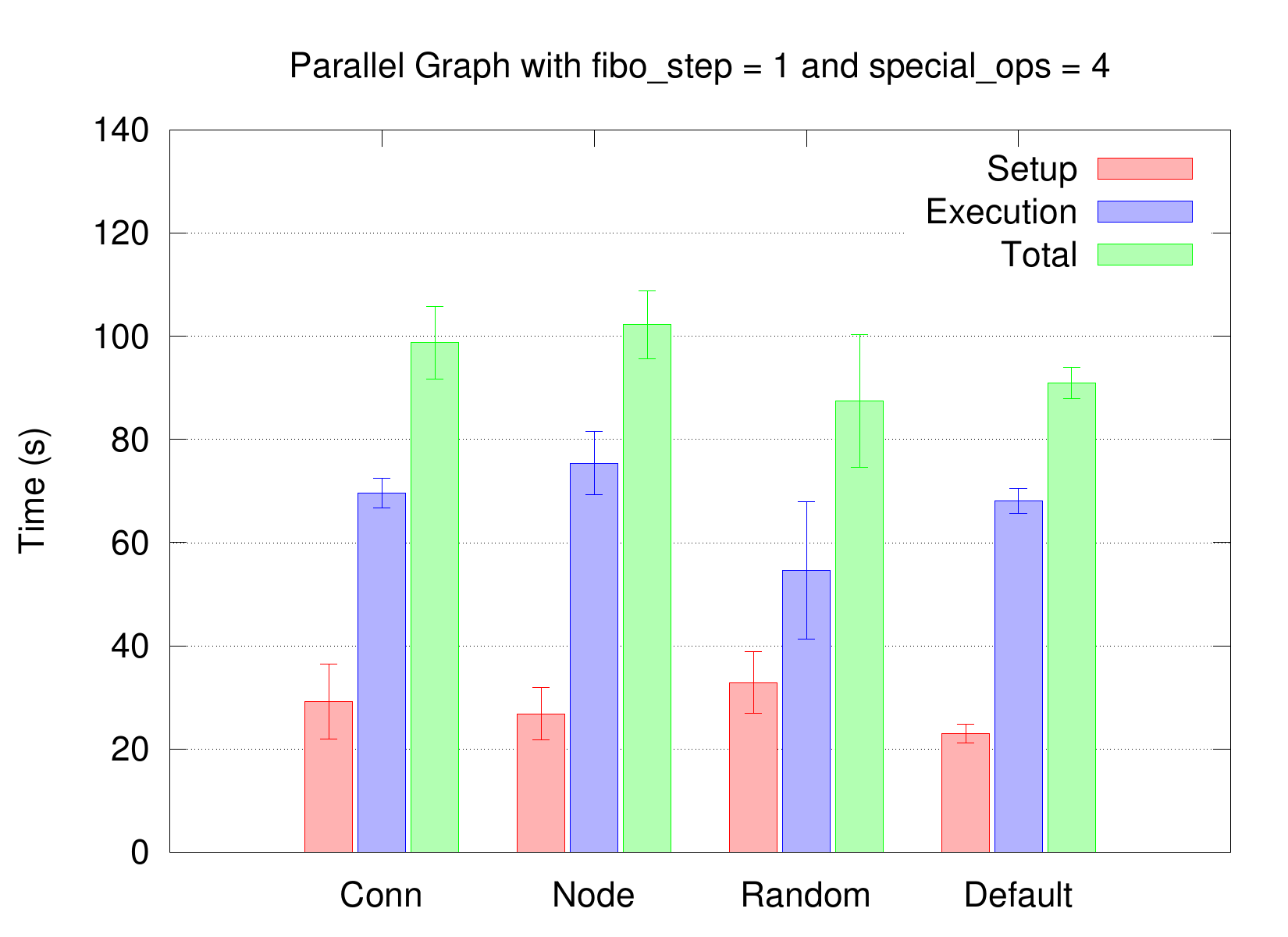}
        \caption{Parallel. Step = 1, sp-ops = 4.}
        \vspace{0.4cm}
        \label{fig:para_st1so3}
    \end{subfigure}
    \begin{subfigure}[b]{0.49\textwidth}
        \centering
        \includegraphics[width=\textwidth]{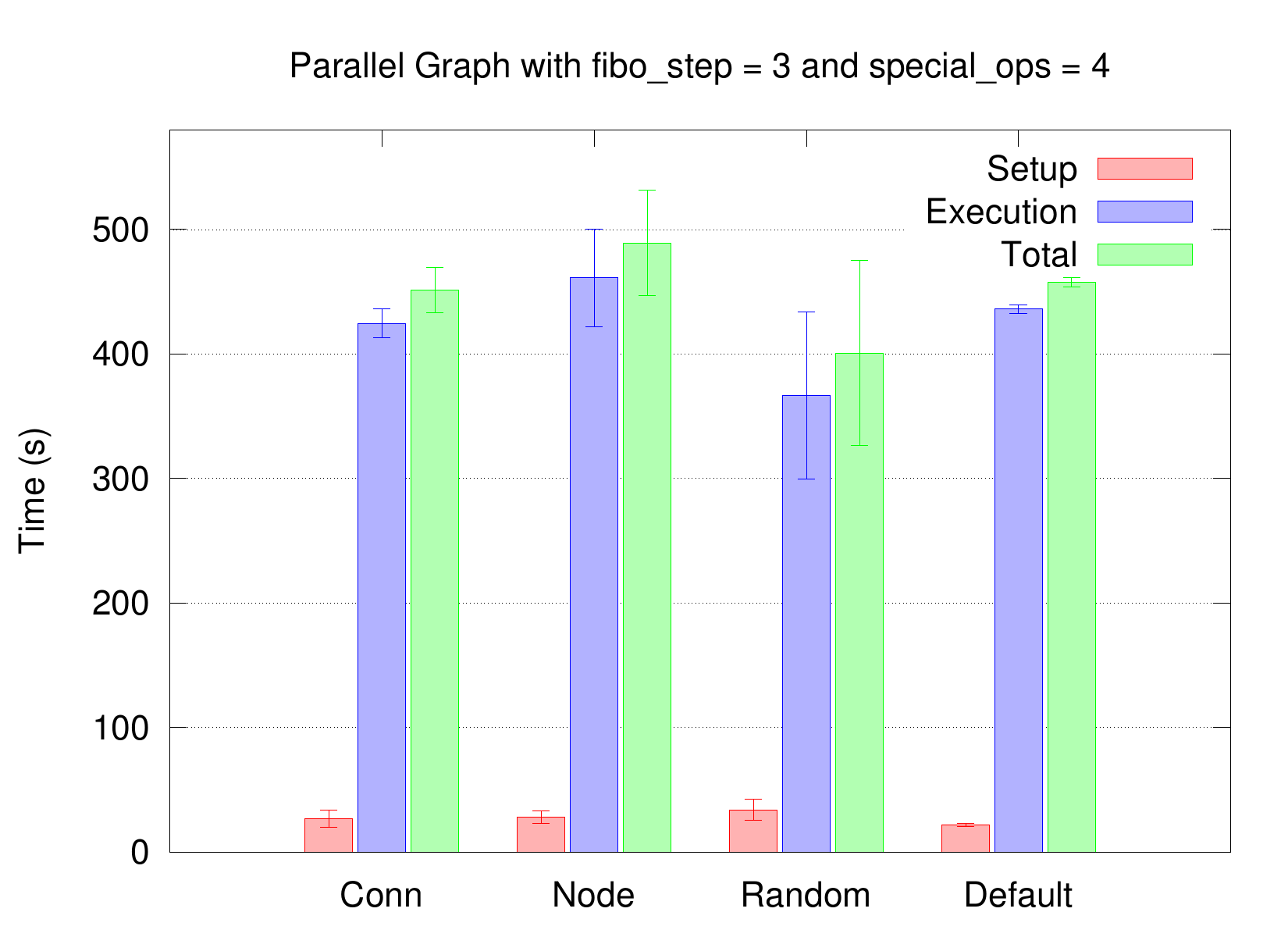}
        \caption{Parallel. Step = 3, sp-ops = 4.}
        \vspace{0.6cm}
        \label{fig:para_st3so3}
    \end{subfigure}
    \caption{Parallel topology experiments}
    \vspace{0.65cm}
    \label{fig:exp2}
\end{figure}

The \textbf{Fibonacci step} is an application parameter related to increasing the $n$ values. 
The $n$ value starts as $5$ (the fifth element of the Fibonacci sequence) and the \textbf{Fibonacci step} ranges from $2$ to $4$. 
As this parameter increases, the overall execution time of the graph increases steeply. 
Figure~\ref{fig:fibo_step} illustrates the propagation of the Fibonacci step parameter in a parallel topology, where $F(n)$ is the algorithm that computes the $n$th element of the sequence, and $fs$ is the \textit{Fibonacci step} parameter. The same logic of an individual line of the parallel topology is applied to the linear topology pipeline.

The pipelines consist of 14 operators, where 12 are computing operators, one is a generator, and one is a terminator. 
The first operator is a \textit{generator}, which sets up the workload for the other operators. 
The generator sends a message to all connected operators, sending the $n$ ($5$ as default) value for the Fibonacci operator. 
This message is sent $5$ times, to ensure continuous communication throughout the execution of the operator. 
The generator is connected with a \textit{terminator} operator, which signals the end of the execution of the pipeline once all other operators finish. 
Finally, the other twelve operators are Fibonacci operators, which receive a message with the $n$ value and compute the $nth$ element of the Fibonacci sequence. 
After computing the sequence, each operator sends a message to the next one with $n + Fibonacci step$, which will set up the execution of the next operator.

\subsection{Experiment Execution}
\label{sec:experimental_framework}

To optimize data pipelines, we set up a framework to measure our experiments, which comprises the following processes:
\begin{itemize}
	\item a parser to convert vFlow graphs into a planning task in PDDL;
	\item a \textbf{numeric planner} to compute a grouping strategy to optimize the parsed pipeline, yielding a PDDL-based plan; and
	\item a converter to interpret this PDDL plan into a vFlow graph, which is executed in an installation of VFLow and extracts performance metrics, e.g., execution time.
\end{itemize} 
VFlow includes a profiling tool to evaluate the performance of the designed approaches, which allows us to collect data on three metrics: \textit{setup time}, \textit{execution time}, and \textit{total time}. 
The \textit{setup time} measures how long it took to prepare the environment; \textit{execution time} tracks how long the pipeline took to execute after the setup phase. 
The \textit{total time} is the setup time plus the execution time.

Figure~\ref{fig:architecture} illustrates the architecture of our approach. 
VFlow exports data pipeline graphs in JSON format, which serves as the medium of communication between VFlow and our optimizer. 
A graph described in JSON is sent to a converter, which converts JSON to a planning description language, PDDL. 
The PDDL has two files, the domain, and problem. 
The domain is where we define the optimization strategy (the ones we detailed in Section \ref{sec:setup}), dictating how we optimize the given pipeline. 
The problem comprises the details of the given graph, detailing the operators (as black boxes) that are going to run, how they are connected forming the topology of the given graph, and the necessary tags to run each operator. 
With these files we execute the planner, ENHSP~\cite{enhsp}, which outputs a plan. 
The plan is the optimization steps, which define the number of groups and which specific images are going to be used to optimized the graph. 
We then use a converter to convert the plan (sequence of optimization steps) to a JSON description of the optimized graph, which runs using the VFlow profiler.

\section{Experimental results}
\label{sec:results}

In the following, we show the quantitative results we obtained from running our experimental evaluation.
We first describe the experimental environment in which we deployed the data pipelines, and follow by a description of the results obtained for each topology of graphs that we generated, i.e. line and parallel.

\subsection{Experimental setup}

The experiments are executed in a virtualized cloud cluster composed of two worker nodes. Docker containers using the Centos-72 image run within the worker nodes, where the containers are organized in pods managed by Kubernetes (version 1.19.13).

To evaluate our approaches, we run each graph five times to account for variations in the cluster's performance. 
We distinguish the measurements of the first execution (cold start) from the other executions (warm starts).
As metrics, we compute the mean setup time, execution time, and total time discussed in the previous section.

We compare our approaches (connection, node, and random) to the \textit{default} baseline that has no grouping assuming that all operators are run in a single group. 
For line topology experiments, it represents an optimum grouping, as all operators are sequential and in the same image.
For parallel topology experiments, the default baseline runs all operators in a single group, which prevents them  to benefit from the parallelism of the topology.
Note that it would be impossible to run the graphs with only one image due if the special tags were real software constraints, so the default baseline is not a viable strategy outside of this evaluation scenario. 
Considering this, we include it as a method for comparison, highlighting the effects of having no groups in different scenarios. 

\subsection{Line topology}

Our first batch of experiments uses the line topology with sequential operators. 
We vary the parameters described in Section~\ref{sec:setup}, running nine distinct baseline graphs, and optimizing each graph using all four approaches, resulting in 36 graphs. 
Figures~\ref{fig:line_st1so1},~\ref{fig:line_st1so3},~\ref{fig:line_st3so3} show the average results in milliseconds for all approaches, dividing these results into \textit{Setup}, \textit{Execution}, and \textit{Total} time. 
Results from these experiments show that the heuristics provide performance gains, mainly in terms of setup time. 
Figures~\ref{fig:line_st1so3} and~\ref{fig:line_st3so3} show that the connection based heuristics also achieved performance gains in execution time. 
Such results indicate that planning heuristics are effective at optimizing stream processing applications.

\subsection{Parallel topology}

The experiments with parallel lines topology use three lines of four operators. 
We vary the parameters described in Section~\ref{sec:setup}, running nine distinct baseline graphs with the same procedure we did for the line topology. 
Figures \ref{fig:para_st1so1}, \ref{fig:para_st1so3}, and \ref{fig:para_st3so3} show the results obtained for three distinct combinations of parameters. 
Figure~\ref{fig:para_st1so3} shows the random approach performing well, even outperforming some heuristics in execution time. The random approach creates more groups, which increases the graph's setup time. Hence, the approaches' contrast in the total time is insignificant if we consider the standard deviation. However, as the execution time is often longer with higher workloads, in some cases, the execution speed-up can compensate the higher time for setup.  

\begin{figure}[t]
    \centering
    \begin{subfigure}[b]{0.49\textwidth}
        \centering
        \includegraphics[width=\textwidth, clip]{graph_of_s1_g1_pdf.pdf}
        \caption{Step = 1, special-ops = 2.}
        \label{fig:para_first1}
    \end{subfigure}
    \begin{subfigure}[b]{0.49\textwidth}
        \centering
        \includegraphics[width=\textwidth, clip]{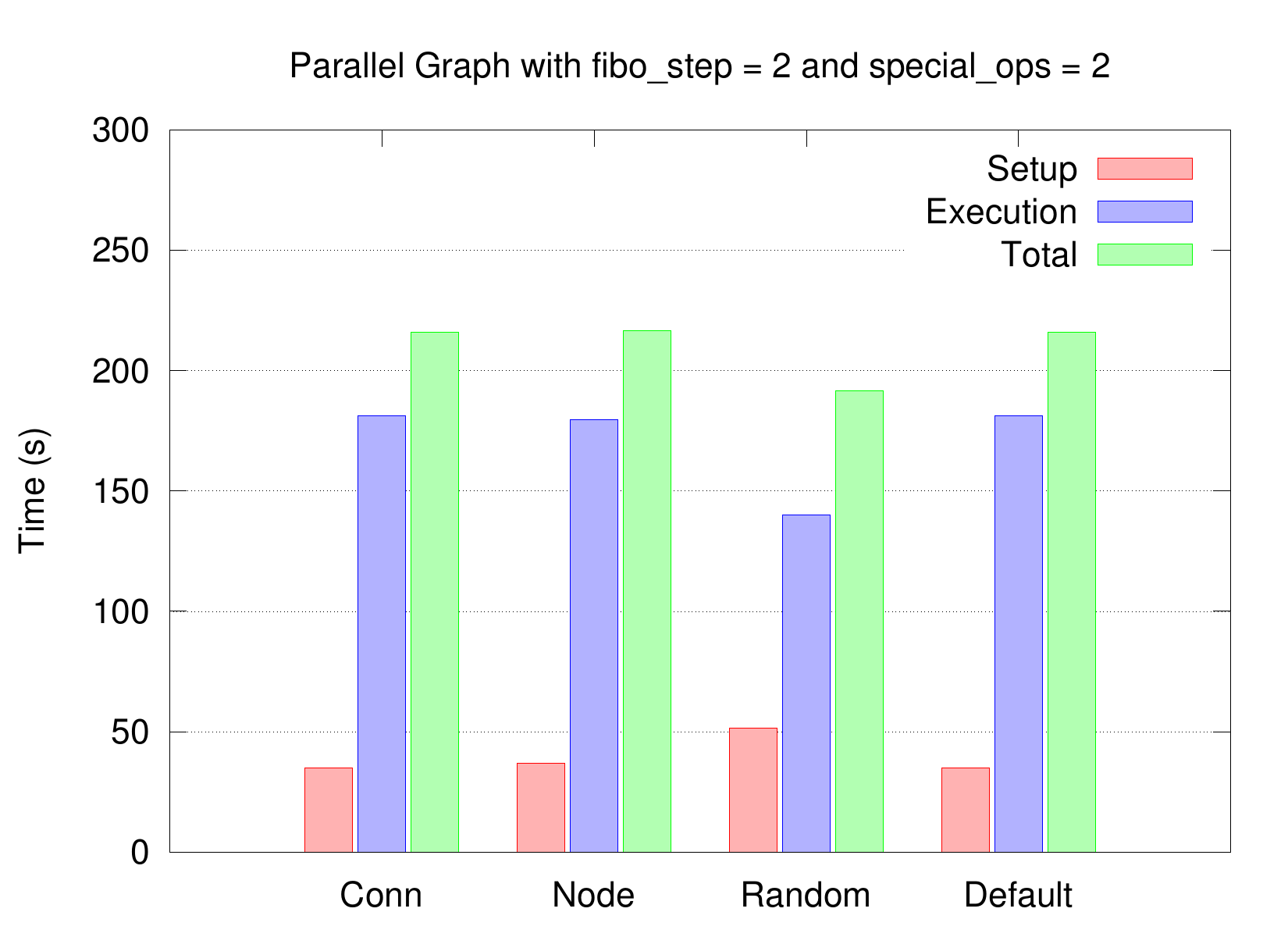}
        \caption{Step = 2, special-ops = 2.}
        \label{fig:para_first2}
    \end{subfigure}
    \vspace{0.6cm}
    \caption{Parallel topology experiments, only first execution.}
    \vspace{0.8cm}
    \label{fig:exp3}
\end{figure}

Figures \ref{fig:para_first1} and \ref{fig:para_first2} show the results for the first execution of two parameters configuration. 
As we can see, the setup time for the first execution is much higher, which impacts the performance of the random approach when the workload is small (Fibonacci step = 1). 
As the workload increases, execution time gains outweigh setup time losses.

\section{Results discussion and closing remarks}
\label{sec:conclusion}

The connection heuristic approach overall had better results. However, although with a high standard deviation, the random strategy with more groups has the best results for the parallel topology. 
We believe that this happens because the random approach created many groups that are capable of executing the processing steps in parallel. Moreover, the results show that more groups for the line topology do not improve the total execution time. The connection strategy, although consistently good in the line topology, does not consider possible parallel executions.

Our results show that the usage of AI techniques can improve the setup and execution time of pipelines by optimizing the grouping of operators.
In this work, our optimization goal was the total execution time. 
In the future, other desirable evaluation metrics can be included, such as the infrastructure cost of running a graph (which includes the number of pods used) or the resilience of the deployed graph.
  
A hybrid strategy could also be explored to optimize each line of the parallel topology individually. 
In future work, we could evaluate grouping strategies based on TCO metrics that measure the total performance time per dollar (on a given cloud platform).

\begin{ack}
We are grateful for the financial support and computing resources from SAP Labs.

\end{ack}



\bibliography{mybibfile}

\end{document}